\documentclass{article}

\usepackage{microtype}
\usepackage{graphicx}
\usepackage{subfigure}
\usepackage{booktabs} 
\usepackage{relsize}
\newcommand{\Acc}[2]{#1 \text{$\pm$\smaller{#2}}}

\usepackage{hyperref}

\usepackage[accepted]{icml2025}
\makeatletter
\renewcommand{\Notice@String}{} 
\makeatother

\usepackage{amsmath}
\usepackage{amssymb}
\usepackage{mathtools}
\usepackage{amsthm}

\usepackage[capitalize,noabbrev]{cleveref}

\theoremstyle{plain}

\theoremstyle{definition}

\theoremstyle{remark}

\usepackage[textsize=tiny]{todonotes}

\begin{document}

\twocolumn[
\icmltitle{Scalable Heterogeneous Graph Learning via Heterogeneous-aware Orthogonal Prototype Experts}




\begin{icmlauthorlist}
\icmlauthor{Wei Zhou}{hust}
\icmlauthor{Hong Huang}{hust}
\icmlauthor{Ruize Shi}{hust}
\icmlauthor{Bang Liu}{ment}
\end{icmlauthorlist}

\icmlaffiliation{hust}{National Engineering Research Center for Big Data Technology and System, Services Computing Technology and System Lab,
Cluster and Grid Computing Lab,
School of Computer Science and Technology,
Huazhong University of Science and Technology, Wuhan,430074, China.}

\icmlaffiliation{ment}{DIRO, Université de Montréal \& Mila \& Canada CIFAR AI Chair, Canada}

\icmlcorrespondingauthor{Hong Huang}{honghuang@hust.edu.cn}
\icmlkeywords{Heterogeneous Graph Neural Networks, }

\vskip 0.3in
]



\printAffiliationsAndNotice{}  

\begin{abstract}
{\it Heterogeneous Graph Neural Networks}~(HGNNs) have advanced mainly through better encoders, yet their decoding/projection stage still relies on a single shared linear head, assuming it can map rich node embeddings to labels. We call this the Linear Projection Bottleneck: in heterogeneous graphs, contextual diversity and long-tail shifts make a global head miss fine semantics, overfit hub nodes, and underserve tail nodes. While {\it Mixture-of-Experts}~(MoE) could help, naively applying it clashes with structural imbalance and risks expert collapse.
We propose a \textbf{H}eterogeneous-aware \textbf{O}rthogonal \textbf{P}rototype \textbf{E}xperts framework named \textbf{HOPE}, a plug-and-play replacement for the standard prediction head. HOPE uses learnable prototype-based routing to assign instances to experts by similarity, letting expert usage follow the natural long-tail distribution, and adds expert orthogonalization to encourage diversity and prevent collapse. Experiments on four real datasets show consistent gains across SOTA HGNN backbones with minimal overhead.
\end{abstract}

\section{Introduction}  
{\it Heterogeneous Graphs}~(HGs) serve as a powerful abstraction for modeling complex systems containing multiple types of nodes and edges. The proliferation of {\it Heterogeneous Graph Neural Networks}~(HGNNs) has significantly advanced the state-of-the-art in tasks such as node classification and link prediction~\cite{SemE, SubInfer}. Pioneering architectures like R-GCN~\cite{R-GCN}, HGT~\cite{HGT}, and the recent scalable SeHGNN~\cite{SeHGNN} and HGAMLP~\cite{HGAMLP} have focused predominantly on the encoding phase, designing intricate mechanisms to aggregate heterogeneous neighbors and fuse multi-modal semantics into a unified node representation.  

However, a critical yet overlooked limitation persists in the decoding or projection phase of these frameworks. The prevailing paradigm implicitly assumes that a single, global linear transformation (followed by Softmax) has sufficient capacity to map the learned high-dimensional representations onto the label space. We identify this assumption as the \textbf{``Fixed Linear Projection Bottleneck''} in HG representation learning.

This bottleneck is particularly severe in HGs due to two intrinsic characteristics. \textbf{1. Contextual Diversity.} A node's semantic meaning is always context-dependent~\cite{metapath2vec, HAN}. For instance, in an academic graph, an author node might behave as a ``theoretician" in one collaboration subgraph but as an ``applied scientist" in another. A static, global projection head forces the encoder to compress these multifaceted roles into a single vector that satisfies an average linear decision boundary, inevitably losing fine-grained semantic nuances. \textbf{2. Long-tail Distribution Shift.} Real-world graphs follow long-tail distributions~\cite{MAG240M, DBLP-PubMed-Yelp}. High-degree ``hub" nodes dominate the gradient updates, causing the global projection matrix to overfit to majority patterns. Consequently, ``tail" nodes, which often require specialized decision boundaries due to their sparse connectivity~\cite{long-tail1, long-tail2}, are poorly served by this global shared head.

{\it Mixture-of-Experts}~(MoE) presents a promising avenue to break this bottleneck. By conditionally activating specialized parameters for different inputs, MoE theoretically allows the model to master diverse semantic patterns without exploding computational costs~\cite{MOE-survey1, MOE-survey2}. However, directly transplanting standard MoE architectures to HGs is non-trivial and fraught with domain-specific challenges.
\textbf{1. Structural Imbalance vs. Load Balancing.}
Standard MoE frameworks typically enforce strict load balancing to ensure all experts are utilized equally~\cite{load-imbalance}. However, HGs inherently exhibit severe long-tail distributions in node types and relation frequencies~\cite{MAG240M, DBLP-PubMed-Yelp}. A ``hub" node appears exponentially more often than a ``tail" type. Forcing a uniform load distribution across experts contradicts this natural data skewness, potentially coercing experts to process node types they are ill-equipped for, thereby degrading performance on both hub and tail classes.
\textbf{2. Expert Collapse.}
Without explicit routing regularization (e.g., load-balancing), MoE models may suffer from routing degeneracy and insufficient expert specialization, leading to redundant experts and reduced benefits of conditional computation~\cite{load-balance1, load-balance2}. In graph settings with heterogeneous and long-tailed semantics, this risk can be exacerbated, motivating additional diversity or balance constraints.

To address these fundamental limitations, we propose a  \textbf{H}eterogeneous-aware \textbf{O}rthogonal \textbf{P}rototype \textbf{E}xperts framework named \textbf{HOPE}. HOPE is explicitly designed to adapt to the heterogeneous information and long-tail semantic distributions inherent in graph-structured data, thereby enhancing the model’s capacity to capture fine-grained, instance-specific patterns.
HOPE introduces a learnable, prototype-based routing mechanism that assigns nodes to experts based on embedding similarity. This approach resolves the conflict between enforced load balancing and the graph's intrinsic long-tail distribution, allowing experts to specialize in distinct semantic regions, including rare patterns.
Furthermore, we incorporate a tailored orthogonalization strategy that enforces orthogonality among dynamic experts. This maximizes semantic diversity and prevents expert collapse, ensuring complementary expertise for complex semantics.
Functioning as a universal plug-and-play module, HOPE can replace the standard prediction head in various HGNN backbones. Extensive experiments on four real-world datasets demonstrate that HOPE consistently enhances the performance of state-of-the-art architectures with minimal computational overhead.

Our contributions can be summarized as follows:
\begin{itemize}
\item \textbf{Universal Projection Paradigm.} We identify the fixed linear head as a bottleneck for polysemous and long-tail nodes. HOPE introduces a model-agnostic hybrid expert layer, serving as a universal plugin to enhance diverse HGNN encoders.
\item \textbf{Orthogonal Experts \& Prototype Routing.} We propose expert orthogonalization to maximize functional diversity and prevent mode collapse, coupled with a prototype-based routing mechanism that mitigates load imbalance by assigning nodes via semantic similarity.
\item \textbf{SOTA Performance \& Compatibility:} Extensive experiments on four datasets confirm that HOPE consistently boosts multiple SOTA backbones with negligible overhead, proving its value as a practical, general-purpose enhancement.
\end{itemize}

\section{Methodology}  
In this section, we first formulate the problem and the limitations of standard projections. We then detail the HOPE architecture, focusing on the prototype routing mechanism and orthogonal constraint.  Finally, we provide a theoretical analysis of the routing properties and a comprehensive complexity analysis.

\subsection{Preliminaries and Motivation}
Let $\mathcal{G} = (\mathcal{V}, \mathcal{E})$ denote a heterogeneous graph. We assume a backbone encoder $\mathcal{F}$ maps each node $v$ to a set of semantic-specific embeddings $\mathcal{H}_v = \{ \mathbf{h}_v^{\Phi_1}, \dots, \mathbf{h}_v^{\Phi_M} \}$, where $\mathbf{h}_v^{\Phi_m} \in \mathbb{R}^d$ corresponds to the $m$-th meta-path view.

\textbf{The Fixed Linear Projection Bottleneck.}
We define the bottleneck as a compound limitation inherent in standard HGNN decoders, arising from the mismatch between rigid model architectures and the dynamic nature of heterogeneous data. It manifests in two aspects:
\begin{itemize}
    \item \textbf{Fixed Capacity Constraint}. Existing heads enforce a uniform computational budget (e.g., a fixed $k$-expert routing or dense aggregation) across all nodes. This ignores the long-tail distribution of node semantics, causing information loss for semantic-rich ``hub" nodes (under-fitting) and introducing noise for sparse ``tail" nodes (over-smoothing).
    \item \textbf{Linear Separability Assumption}. The reliance on a global linear projection implicitly assumes that complex node embeddings are linearly separable. This fails to capture the severe contextual polysemy and distribution shifts, limiting the model's ability to resolve non-linear decision boundaries in the latent space.
\end{itemize}

\subsection{Overview of HOPE Framework}
\begin{figure*}[t]
    \centering
    \includegraphics[width = 0.75\textwidth]{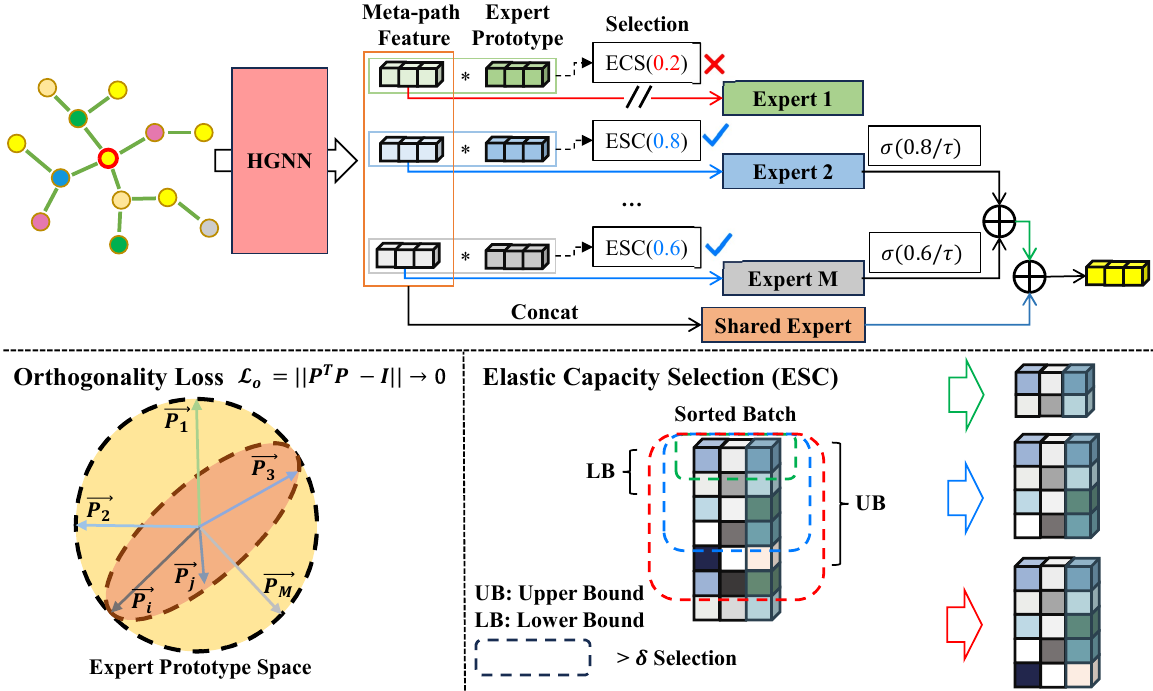}
    \caption{The structure of HOPE.}
    \label{fig:framework}
\end{figure*}
To overcome the limitations of linear projections, we propose \textbf{HOPE} (\textbf{H}eterogeneous-aware \textbf{O}rthogonal \textbf{P}rototype \textbf{E}xperts). 
As shown in Figure~\ref{fig:framework}, HOPE is a plug-and-play neural projection framework that transforms the traditional decoding layer into a MoE system. It accepts representations from any HGNN backbone and refines them to handle inherent long-tail distributions. Instead of strict load balancing, HOPE employs a **prototype routing mechanism**, selecting experts based on semantic similarity to learnable prototypes to naturally adapt to data skewness. Furthermore, an **elastic capacity mechanism** dynamically adjusts the number of active experts based on feature quality under a global budget.
Architecturally, the projection is decoupled into two independent pathways:
1. Shared Pathway. Captures stable, universal structural patterns across the graph.
2. Dynamic Pathway. Sparsely activated to handle instance-specific semantic nuances.
To prevent redundancy and mode collapse, we enforce strict orthogonality among dynamic experts. By replacing standard prediction heads, HOPE effectively unlocks the backbone's potential in modeling complex heterogeneity with minimal computational overhead.

\subsection{Input and Context Encoding}
Given a heterogeneous graph $\mathcal{G}=(\mathcal{V}, \mathcal{E})$ and a set of predefined meta-paths $\mathcal{M} = \{\Phi_1, \Phi_2, \dots, \Phi_M\}$, we assume a backbone encoder $\mathcal{F}_{\theta}$ produces a set of semantic-specific embeddings for each node $v$
\begin{equation}
    \mathcal{H}_v = \{ \mathbf{h}_v^{\Phi_1}, \mathbf{h}_v^{\Phi_2}, \dots, \mathbf{h}_v^{\Phi_M} \},
\end{equation}
where $\mathbf{h}_v^{\Phi_m} \in \mathbb{R}^d$ captures the node semantics under the $m$-th meta-path view.
To provide a holistic view for the global stability analysis, we also construct a concatenated representation
\begin{equation}
    \mathbf{h}_v^{\text{all}} = \text{Concat}(\mathbf{h}_v^{\Phi_1}, \dots, \mathbf{h}_v^{\Phi_M}) \in \mathbb{R}^{M \cdot d}.
\end{equation}

\subsection{Hybrid Expert Mechanism}
HOPE employs a hybrid architecture where experts are explicitly aligned with meta-path views to balance global structural stability and local semantic selection.

\subsubsection{Shared Pathway}
To capture cross-view dependencies and universal patterns by observing the complete semantic context simultaneously.
We deploy a shared expert $S$. The input to this pathway is the concatenated feature vector $\mathbf{h}_v^{\text{all}}$. The stability representation $\mathbf{z}_{v}^{(s)}$ is computed as
\begin{equation}
    \mathbf{z}_{v}^{(s)} = S(\mathbf{h}_v^{\text{all}}),
\end{equation}
where $S$ is a {\it Multilayer Perceptron}~(MLP). By processing the full spectrum of meta-paths, the shared expert ensure that the model retains a comprehensive understanding of the node's global context.

\subsubsection{Dynamic Pathway}
To adaptively select and refine the most discriminative meta-path views for each specific node instance.
Distinct from generic MoEs, we design the dynamic expert bank to strictly correspond to the meta-path set. We instantiate $M$ dynamic experts $\{D_1, \dots, D_M\}$, where the $m$-th expert $D_m$ is dedicated to processing the specific feature $\mathbf{h}_v^{\Phi_m}$.

{\bf Semantic-Aware Gating}. Each expert $D_m$ is associated with a learnable prototype $\mathbf{p}_m \in \mathbb{R}^d$, representing the ideal semantic signature for that meta-path view.
Instead of a global routing competition, the activation of expert $m$ is determined by the alignment between its specific input $\mathbf{h}_v^{\Phi_m}$ and its prototype $\mathbf{p}_m$. The relevance score is
\begin{equation}
    s_{v,m} = \frac{(\mathbf{h}_v^{\Phi_m})^\top \mathbf{p}_m}{\|\mathbf{h}_v^{\Phi_m}\|_2 \|\mathbf{p}_m\|_2}.
\end{equation}

{\bf Elastic Capacity Selection}. 
To balance elastic computational resource allocation while preventing expert collapse, we propose a hybrid expert selection mechanism. This mechanism addresses the conflict between signal noise interference and tail node loss in traditional routing strategies through three distinct levels of constraints:

\begin{itemize}
    \item \textbf{Quality Criterion ($\mathcal{C}_{Qual}$) --- Noise Rejection}.
    To ensure experts only process high-confidence samples and avoid interference from irrelevant signals during feature learning, we set a hard threshold $\delta$, considering only nodes with semantic alignment $s_{v,m} > \delta$. This constitutes the primary filtering barrier.

    \item \textbf{Stability Criterion ($\mathcal{C}_{Stab}$) --- Tail Preservation}.
    Targeting ``long-tail'' nodes with weak feature representations that are easily filtered out by thresholds (i.e., hard samples), the expert enforces the inclusion of the Top-$K$ best-matching nodes in the batch (as a lower bound) to prevent them from becoming unassigned ``orphan nodes'' and causing information loss. This provides a survival guarantee mechanism.

    \item \textbf{Capacity Criterion ($\mathcal{C}_{Cap}$) --- Load Control}.
    Constrained by the computational budget, to prevent specific experts from overloading and to maintain system efficiency, we ultimately retain only the highest-scoring Top-$C$ nodes (as an upper bound) from the union of the above two criteria. This achieves memory-friendly elastic truncation.
\end{itemize}

Formally, the selection mask $M_{v,m} \in \{0, 1\}$ is defined as
\begin{equation}
    M_{v,m} = 
    \begin{cases} 
    1 & \text{if } v \in \operatorname{Top-{\it C}}(\Omega_m ) \\
    0 & \text{otherwise}
    \end{cases},
\end{equation}
where $\Omega_m = \mathcal{S}_{Qual}^{(m)} \cup \mathcal{S}_{Stab}^{(m)}$ denotes the candidate pool, $\mathcal{S}_{Qual}^{(m)}=\{ u \mid s_{u,m} > \delta \}$ represents the high-confidence node set filtered by semantic alignment, and $\mathcal{S}_{Stab}^{(m)}=\operatorname{Top-{\it K}}( \mathbf{s}_{:,m})$ is the mandatory set of the top-$K$ best-matching nodes serving as a stability guarantee.


\textbf{Gated Aggregation}.
For the selected node-expert pairs, we compute a gating weight using a Sigmoid function with temperature $\tau$ as 
\begin{equation}
    g_{v,m} = \sigma(s_{v,m} / \tau) \cdot M_{v,m},
\end{equation}
where $\tau$ is a trainable parameter.
The dynamic representation $\mathbf{z}_{v}^{(d)}$ is the sparse weighted sum of activated experts
\begin{equation}
    \mathbf{z}_{v}^{(d)} = \sum_{m=1}^M g_{v,m} \cdot D_m(\text{LayerNorm}(\mathbf{h}_v^{\Phi_m})).
\end{equation}

\subsubsection{Adaptive Fusion}
The final node representation $\mathbf{z}_v$ fuses the global context from the shared pathway and the selected semantics from the dynamic pathway via a projection matrix $\mathbf{W}_{\text{fusion}}$ as 
\begin{equation}
    \mathbf{z}_v = \mathbf{W}_{\text{fusion}} \big[ \mathbf{z}_{v}^{(s)} \mathbin{\|} \mathbf{z}_{v}^{(d)} \big] + \mathbf{b}.
\end{equation}

\subsection{Orthogonality Constraint}
To ensure that different meta-path experts represent distinct semantic directions. 
Even though experts are tied to different meta-paths, their prototypes should remain distinct to prevent semantic overlap. We enforce orthogonality on the prototype matrix $\mathbf{P} = [\mathbf{p}_1, \dots, \mathbf{p}_M]$ as 
\begin{equation}
    \mathcal{L}_{\text{o}} = \| \mathbf{P}^\top \mathbf{P} - \mathbf{I} \|_F^2 = \sum_{i \neq j} \left( \frac{\mathbf{p}_i^\top \mathbf{p}_j}{\|\mathbf{p}_i\|_2 \|\mathbf{p}_j\|_2} \right)^2.
\end{equation}

The total objective function is:
\begin{equation}
    \mathcal{L} = \mathcal{L}_t + \lambda \mathcal{L}_{\text{ortho}},
\end{equation}
where $\mathcal{L}_t$ denotes the target loss of downstream tasks.

\subsection{Theoretical Analysis}
\label{sec:theory}
We analyze the effectiveness of the proposed Elastic Capacity Selection mechanism from the perspectives of gradient flow preservation and noise robustness.

\textbf{Robustness against Representation Collapse}.
\label{prop:collapse}
Consider a scenario where a specific meta-path view $\Phi_{noise}$ generates non-informative features for a subset of nodes $\mathcal{V}_{sub}$, resulting in alignment scores $s_{v, \Phi_{noise}} \approx 0, \forall v \in \mathcal{V}_{sub}$.
In standard Top-$k$ routing (where $k$ is fixed), if other experts are saturated, the noisy expert $D_{noise}$ is forced to process $k$ nodes, introducing variance $\sigma^2_{noise}$ into the optimization landscape.
In HOPE, the Quality Criterion ($\mathcal{C}_{Qual}$) with threshold $\delta$ ensures that if $\max_{v \in \mathcal{B}} s_{v, \Phi_{noise}} < \delta$, the active set size approaches the lower bound $|\Omega_{noise}| \to K$. Since $K \ll C$ (typically $K=1$), the injection of noise is strictly bounded, minimizing the interference term in the gradient update $\nabla_{\theta} \mathcal{L}$.

\textbf{Gradient Preservation for Tail Nodes}.
\label{prop:gradient}
Let $v_{tail}$ be a hard sample with weak semantic features such that its maximum alignment score $\max_m s_{v_{tail}, m} < \delta$. Under a pure thresholding strategy (e.g., vanilla sparse gating), the gating weights would be zeroed out ($g_{v_{tail}, m} = 0, \forall m$), leading to the ``dead node'' problem where $\frac{\partial \mathcal{L}}{\partial \mathbf{h}_{v_{tail}}} \to \mathbf{0}$.
The Stability Criterion ($\mathcal{C}_{Stab}$) guarantees that $v_{tail}$ is included in the candidate set $\Omega_{m^*}$ of its relatively best-matching expert $m^* = \arg\max_m s_{v_{tail}, m}$. This ensures a non-zero gradient flow path back to the backbone encoder, preventing the feature learning stagnation often observed in long-tail distributions.

\textbf{Semantic Disentanglement}.
The orthogonality constraint on prototypes $\mathcal{L}_{\text{o}}$ minimizes the cosine similarity between expert directions. This effectively pushes the expert prototypes $\mathbf{p}_m$ to span the maximal volume in the feature space $\mathbb{R}^d$. Consequently, the Semantic-Aware Gating becomes a projection onto a nearly orthogonal basis, maximizing the discriminative power of the routing mechanism compared to random initialization or dense attention mechanisms.

\subsection{Complexity Analysis}
\label{sec:complexity}
We demonstrate that HOPE achieves superior representational capacity with a computational cost comparable to standard GNNs, maintaining linear scalability with respect to batch size.

\paragraph{Time Complexity.}
Let $|\mathcal{B}|$ be the batch size, $d$ the hidden dimension, $M$ the number of meta-paths, and $L$ the number of layers in each MLP expert.
\begin{itemize}
    \item \textbf{Backbone \& Pre-processing:} Generating meta-path embeddings takes similar time to exist HGNN models.
    
    \item \textbf{Shared Pathway:} The shared expert $S$ processes the concatenated input for all nodes. Since the input dimension is $M \cdot d$, the first layer costs $O(|\mathcal{B}| \cdot (M \cdot d) \cdot d)$, and subsequent $L-1$ layers cost $O(|\mathcal{B}| \cdot d^2)$. Thus, the total cost is $O(|\mathcal{B}| \cdot d^2 \cdot (M + L))$.
    
    \item \textbf{Routing Decision:} Computing alignment scores $s_{v,m}$ requires dot products between node features and prototypes: $O(|\mathcal{B}| \cdot M \cdot d)$. The selection process (Top-$K$ and filtering) is efficient on GPUs, typically dominated by the score computation.
    
    \item \textbf{Dynamic Execution (Sparse):}
    Crucially, experts in the dynamic pathway only process a subset of nodes. Let $\rho$ be the average sparsity ratio (active nodes / total nodes) determined by the Elastic Capacity Selection ($\rho \ll 1$). Each dynamic expert is an $L$-layer MLP. The computation cost is $O(\rho \cdot |\mathcal{B}| \cdot M \cdot L \cdot d^2)$.
\end{itemize}
The overall time complexity is roughly $O(|\mathcal{B}| \cdot d^2 \cdot (M + \rho M L))$. Since $\rho M \approx \text{const}$ (average active experts per node) and $L$ is small, HOPE is asymptotically linear with respect to $|\mathcal{B}|$.

\paragraph{Space Complexity.}
The parameter overhead is minimal compared to the performance gain:
\begin{itemize}
    \item \textbf{Model Parameters:} We store $M$ dynamic experts and 1 shared expert. Assuming each expert has $L$ layers of width $d$, the parameters are approximately $O((M+1) \cdot L \cdot d^2)$. Since $M$ is typically small (e.g., $M < 10$) in HG datasets, this is negligible compared to large-scale graph adjacency matrices.
    \item \textbf{Routing Prototypes:} Storing $M$ prototype vectors requires only $O(M \cdot d)$.
    \item \textbf{Memory Usage:} The gating mechanism utilizes a sparse selection mask, avoiding the storage of a dense $|\mathcal{B}| \times M$ intermediate activation tensor, making HOPE memory-efficient for large batches.
\end{itemize}

\section{Experiments}
\label{sec:experiments}

In this section, we rigorously evaluate HOPE on several real-world datasets. We aim to answer the following questions:
\begin{itemize}
\item \textbf{RQ1 (Performance \& Efficiency):} Can the proposed HOPE architecture help HGNN models perform better with acceptable additional overhead?
\item \textbf{RQ2 (Ablation Study):} Are the shared pathway, Orthogonality constraints, and elastic capacity selection necessary for performance?
\item \textbf{RQ3 (Parameter Sensitivity):} How do the similarity threshold $\delta$, lower bound $K$, upper bound $C$, and orthogonality constraint weight $\lambda$ affect model behavior?
\end{itemize}

\subsection{Experimental Setup}
\paragraph{Datasets.}
To demonstrate the performance and efficiency of HOPE, a general-purpose performance enhancement strategy, we evaluate it on two types of tasks: node classification and link prediction. The node classification task includes three heterogeneous graphs (Freebase~\cite{Freebase}, Ogbn-mag~\cite{MAG240M}, and DBLP~\cite{DBLP-PubMed-Yelp}). The link prediction task includes two heterogeneous graphs (Yelp~\cite{DBLP-PubMed-Yelp} and DBLP). Detailed information about the data is presented in  Table~\ref{tab:dataset}.
\begin{table}[t]
\centering
\caption{Dataset statistics.}
\label{tab:dataset}
\begin{tabular}{c|c|c|c}
\toprule
Dataset & \#Node & \#Edge & Task \\
\midrule
Freebase & 180,098 & 2,115,376 & NC \\
\midrule
Ogbn-mag & 1,939,743 & 42,222,014 & NC \\
\midrule
DBLP & 1,989,077 & 376,486,176 & NC/LP \\
\midrule
Yelp & 8,2465 & 30,542,675 & LP\\
\bottomrule
\end{tabular}
\end{table}

\paragraph{Baselines.}
To evaluate the effectiveness of HOPE, we conduct a comparative analysis of HOPE against existing enhancement techniques. This comparison occurs on two categories of backbone models: 1) traditional methods, such as R-GCN~\cite{R-GCN}, R-GAT~\cite{GAT}, and R-GSN~\cite{R-GSN}. 2) state-of-the-art methods that employ specific enhancement techniques, including NARS~\cite{NARS}, SeHGNN~\cite{SeHGNN}, and HGAMLP~\cite{HGAMLP}. 

\paragraph{Experimental Settings.}
All models are trained on the server equipped with the Intel Xeon CPU E5-2680, the Nvidia RTX 5090 (32GB) GPU, and 50GB of memory. The software used was Ubuntu 22.04 and CUDA 11.8.  
All models are implemented in PyTorch with DGL backend. 
We use Adam optimizer with learning rate 0.001 and weight decay 0.0. All models are trained for 300 epochs. 

For R-GCN, R-GAT, and R-GSN, the hidden dimension is set to 256, with 2-layer architectures. The representation of each hop is passed through an $mlp$, and the average of all layers is then used as the final node representation. When enhanced with HOPE, the input also consists of the representations from each hop. The similarity threshold $\delta$ is set to 0.2, lower bound $K$ to 0.5, upper bound $C$ to 3, and the orthogonality constraint weight $\lambda$ to 0.5. Due to the excessive time complexity of these methods, we use neighbor sampling to reduce time overhead. For all datasets, the number of sampled neighbors is 15 for the first hop and 20 for the second hop. 

For NARS, SeHGNN, and HGAMLP, the hidden dimension is set to 512, with 2-layer architectures. When enhanced with HOPE, the input for NARS is the node representations of each heterogeneous subgraph, while the inputs for SeHGNN and HGAMLP are the representations corresponding to each meta-path. The similarity threshold $\delta$ is set to 0.6, lower bound $K$ to 0.5, upper bound $C$ to 3, and the orthogonality constraint weight $\lambda$ to 0.5.

Except for Ogbn-mag, the node representations for all datasets are generated by ComplEx~\cite{complex}, whereas in Ogbn-mag, the representations for nodes other than papers are generated by LINE~\cite{Line}.
The other parameters not mentioned are set to their default values. To obtain accurate results, we repeat all experiment 10 times and take the average~(random seeds are $[1, 2, 3, 4, 5, 6, 7, 8, 9, 10]$).

\subsection{Node Classification}
\begin{table*}[t]
\centering
\caption{Node classification results. OOM denotes Out of Memory, and table shows the time spent in one epoch.
}
\label{tab:experiment-nc}
\fontsize{9pt}{4pt}\selectfont
\setlength{\tabcolsep}{2pt}
{
\begin{tabular}{c|cccc|cccc|cccc}
\toprule
& \multicolumn{4}{c|}{Freebase} & \multicolumn{4}{c|}{Ogbn-mag} & \multicolumn{4}{c}{DBLP}\\
\midrule
 & \#Parma & Training & Inferring & Accuracy & \#Parma & Training & Inferring & Accuracy & \#Parma & Training & Inferring & Accuracy\\
\midrule[1pt]
R-GCN & 0.12M & 0.22s & 6.83s & \Acc{61.19}{0.22} & 0.19M & 103.62s & 18.22s & \Acc{35.34}{0.61} & 0.13M & 4.67s & 6.79s & \Acc{42.17}{2.47}\\
\midrule
w. HOPE & 0.15M & 0.17s & 6.90s & \Acc{63.02}{0.14} & 0.21M & 108.11s & 20.03s & \Acc{38.43}{0.52} & 0.16M & 5.24s & 5.83s & \Acc{53.56}{0.49}\\
\midrule[1pt]
R-GAT & 0.12M & 0.25s & 7.34s & \Acc{62.06}{0.25} & 0.19M & 132.81s & 21.83s & \Acc{34.28}{0.17} & 0.13M & 4.46s & 6.39s & \Acc{43.30}{0.49}\\
\midrule
w. HOPE & 0.15M & 0.24s & 7.32s & \Acc{63.97}{0.23} & 0.24M & 109.64s & 21.36s & \Acc{38.97}{0.37} & 0.16M & 5.47s & 5.88s & \Acc{55.27}{2.47}\\
\midrule[1pt]
R-GSN & 0.17M & 0.35s & 8.60s & \Acc{63.98}{1.02} & \multicolumn{4}{c|}{OOM} & 0.18M & 4.84s & 6.65s & \Acc{53.28}{2.47}\\
\midrule
w. HOPE & 0.18M & 0.36s & 8.54s & \Acc{65.55}{0.82} & \multicolumn{4}{c|}{OOM} & 0.19M & 5.62s & 6.54s & \Acc{58.12}{2.26}\\
\midrule[1pt]
NARS & 4.0M & 0.04s & 0.65s & \Acc{66.97}{0.17} & 4.1M & 8.12s & 11.11s & \Acc{52.49}{0.22} & 4.1M & 0.01s & 28.67s & \Acc{54.99}{0.99}\\
\midrule
w. HOPE & 4.3M & 0.05s & 0.69s & \Acc{68.33}{0.19} & 4.6M & 9.01s & 12.74s & \Acc{56.87}{0.35} & 4.6M & 0.02s & 30.19s & \Acc{57.03}{0.83}\\
\midrule[1pt]
SeHGNN & 56.5M & 0.12s & 0.62s & \Acc{64.88}{0.16} & 9.26M & 7.57s & 0.46s & \Acc{57.76}{0.13} & 10.1M & 0.02s & 15.62s & \Acc{50.14}{1.45}\\
\midrule
w. HOPE & 75.9M & 0.15s & 0.63s & \Acc{66.73}{0.52} & 11.9M & 8.21s & 0.46s & \Acc{57.95}{0.14} & 13.6M & 0.02s & 15.96s & \Acc{51.28}{0.69}\\
\midrule[1pt]
HGAMLP & 56.1M & 0.18s & 0.53s & \Acc{65.67}{0.57} & 8.86M & 5.78s & 0.45s & \Acc{57.89}{0.12} & 9.66M & 0.02s & 11.16s & \Acc{49.86}{1.07}\\
\midrule
w. HOPE & 75.6M & 0.20s & 0.56s & \Acc{67.90}{0.13} & 11.5M & 6.76s & 0.43s & \Acc{58.01}{0.12} & 13.3M & 0.02s & 11.04s & \Acc{50.47}{0.40}\\
\bottomrule
\end{tabular}}
\end{table*}
To comprehensively evaluate the generality and core advantages of the HOPE architecture, we conduct extensive node classification experiments on three real-world datasets covering varying scales and topological characteristics. The experimental results are shown in Table \ref{tab:experiment-nc}. The experimental results not only fully validate the effectiveness of HOPE in handling diverse data but also reveal the exceptional balance achieved between computational efficiency and model performance. Notably, HOPE significantly enhances the overall classification performance of base models while introducing negligible additional computational overhead, and even demonstrating higher inference efficiency in certain optimized scenarios.

Specifically, in the most challenging experimental settings, HOPE achieves a performance leap compared to baseline models. This significant quantitative metric strongly confirms that the architecture possesses exceptional robustness and superior scalability when facing complex real-world application scenarios.

Further in-depth analysis indicates that the key to HOPE's success lies in its fundamental overcoming of the ``linear constraint bottleneck" prevalent in traditional methods. Traditional models are often limited by shallow linear expressive capabilities, making it difficult to capture deep interaction features within the data. In contrast, by introducing more expressive non-linear mechanisms, HOPE successfully achieves effective modeling and precise capturing of the implicit, complex non-linear semantic information within the data. This capability enables the model to maintain stable and substantial gains regardless of drastic changes in data scale or severe distribution shifts.

\subsection{Ablation Studies}
\begin{table}[t]
\centering
\caption{Ablation Study on Ogbn-mag with backbone HGAMLP.}
\label{tab:ablation}
\begin{tabular}{lccc}
\toprule
\textbf{Variant} & Train & Infer & Acc \\
\midrule
\textbf{HOPE (Full Model)} & 6.76s & 0.43s & 58.01 \\
w/o Shared Pathway & 6.24s & 0.30s & 57.13 \\
w/o Prototype Experts & 5.98s & 0.28s & 55.36 \\
w/o Elastic Capacity & 6.04s & 0.35s & 57.44 \\
w/o $\mathcal{L}_{\text{o}}$& 6.63s & 0.41s & 57.38 \\
\bottomrule
\end{tabular}
\end{table}
To validate the necessity of each component in the HOPE architecture, we conduct a comprehensive ablation study on the Ogbn-mag dataset using HGAMLP as the backbone model. The relevant results are summarized in Table~\ref{tab:ablation}.

\textbf{Impact of Shared Pathway}. Removing the shared pathway results in a significant performance drop. This result strongly confirms the critical role of the shared expert as an ``anchor". By providing a stable global context representation, it effectively mitigates uncertainty in the routing process, enabling dynamic experts to focus on capturing specific local patterns without the extra burden of maintaining global structural information.

\textbf{Impact of Prototype Experts}. In this setting, we remove the expert prototypes and instead use an $mlp$ to assign nodes to experts. Each node selects the top-$2$ experts. Additionally, we incorporate a load balancing loss with a weight of 0.5 to ensure that no experts are starved. The results show that replacing prototype-based routing with a standard $mlp$ router leads to a noticeable drop in performance. This suggests that the semantic guidance provided by the prototypes is crucial for accurate expert assignment. The $mlp$ router, despite the load balancing loss, struggles to effectively cluster semantically similar nodes to the appropriate experts, resulting in suboptimal feature specialization and reduced overall classification accuracy.

\textbf{Impact of Elastic Capacity}. In this setting, we remove the lower and upper bounds on expert capacity. We observe that this setting leads to the most significant performance degradation, highlighting the importance of the ``leave no node behind" principle. For nodes with weak features or those in the long-tail distribution, forcibly assigning experts (even with slightly lower matching scores) ensures effective gradient backpropagation. This prevents them from becoming ``dead nodes" that cannot be updated, thereby guaranteeing the integrity of representation learning. Moreover, removing the upper bound results in a significant increase in computational overhead. The upper bound prevents ``hub nodes'' from monopolizing experts, ensuring load balance and efficiency.

\textbf{Impact of Orthogonality}. It is worth noting that removing the prototype orthogonal loss ($L_{o}$) results in the most drastic performance loss. This phenomenon profoundly reveals the core status of the orthogonality constraint: it forces experts to maintain differentiation in the feature space, preventing the problem of expert homogenization (Collapse). By maximizing the mutual exclusivity between experts, this mechanism ensures that the model can capture diverse and complementary semantic features in the data, rather than repeatedly learning similar patterns.

\subsection{Parameter Sensitivity Analysis}
\begin{figure}[t]
\centering
\subfigure[$\lambda$]
{
\includegraphics[width=0.23\textwidth]{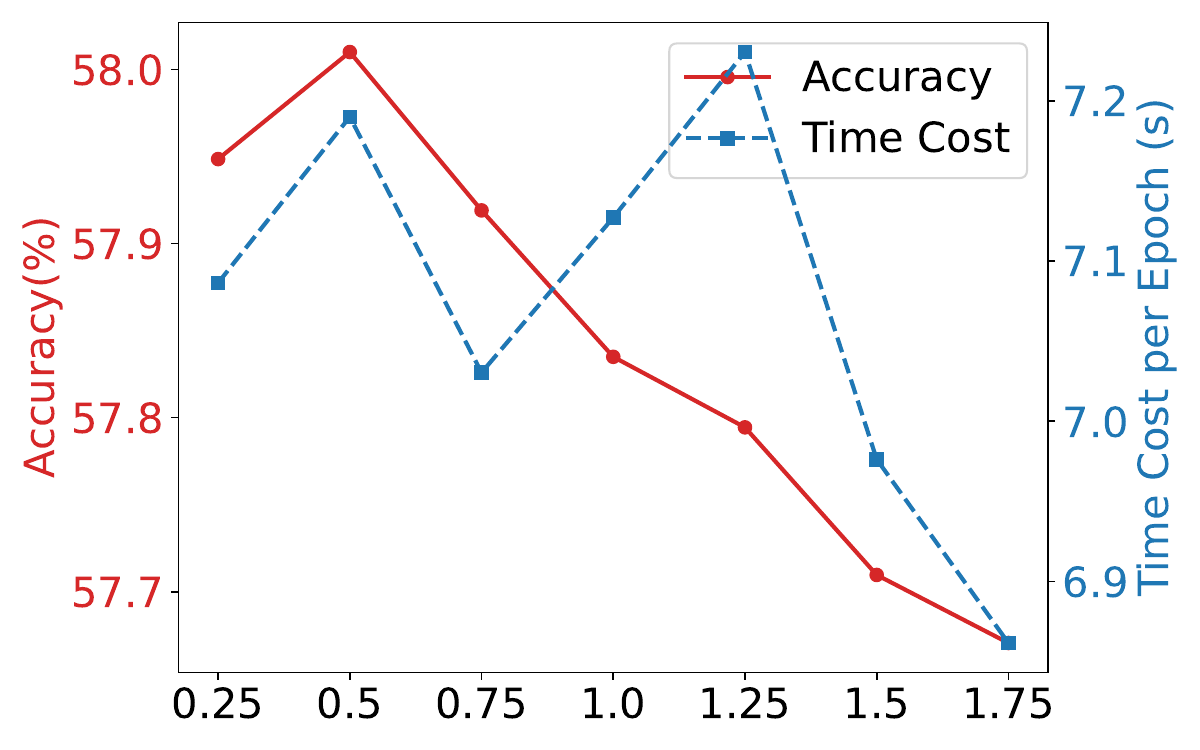}
}
\subfigure[$\sigma$]
{
\includegraphics[width=0.22\textwidth]{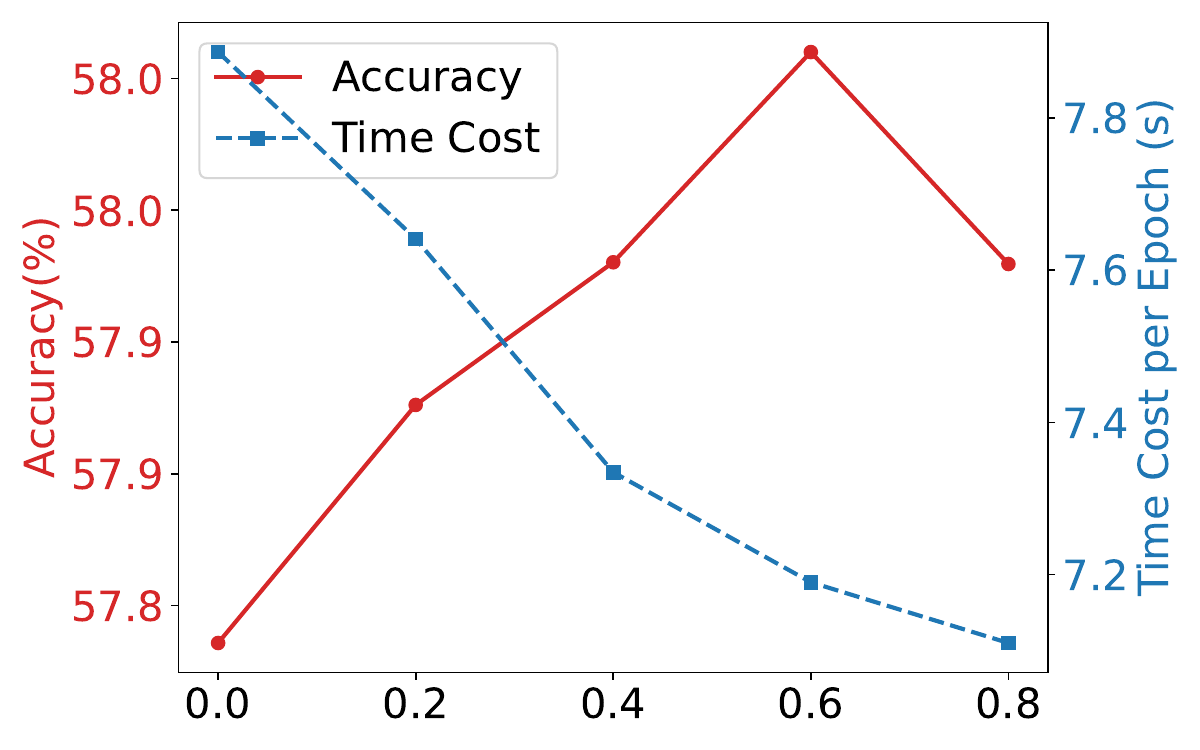}
}\\
\subfigure[$C$]
{
\includegraphics[width=0.23\textwidth]{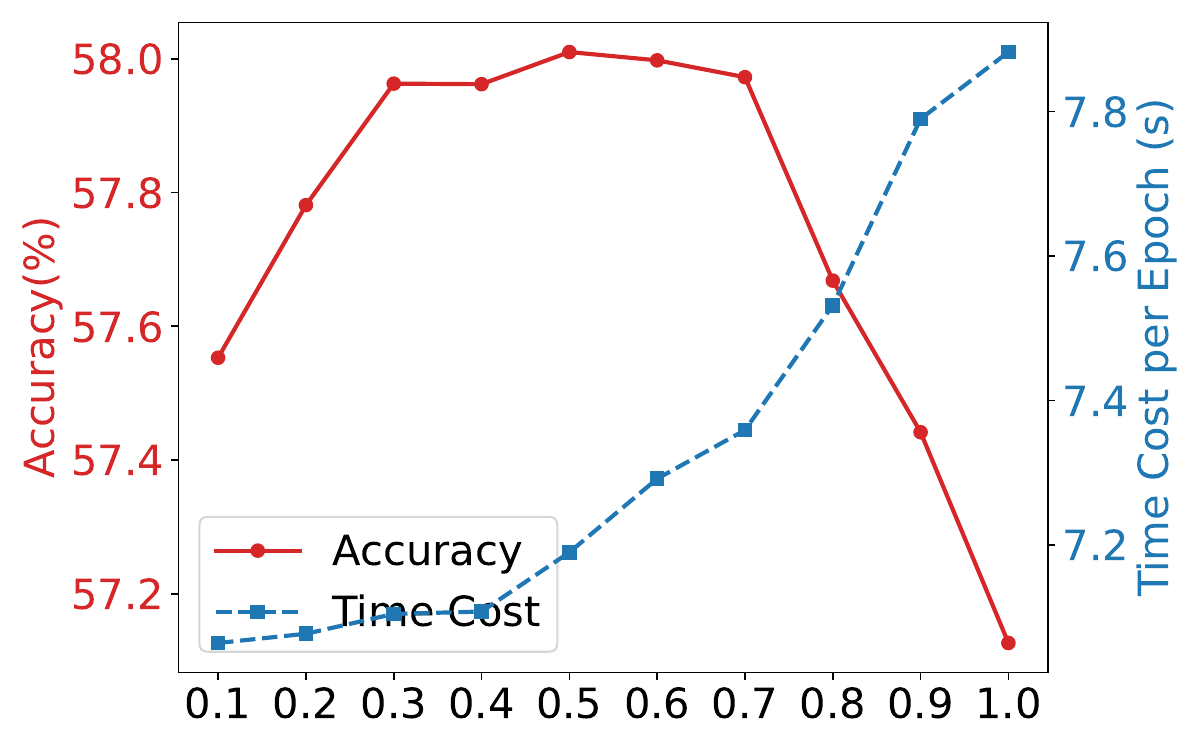}
}
\subfigure[$K$]
{
\includegraphics[width=0.22\textwidth]{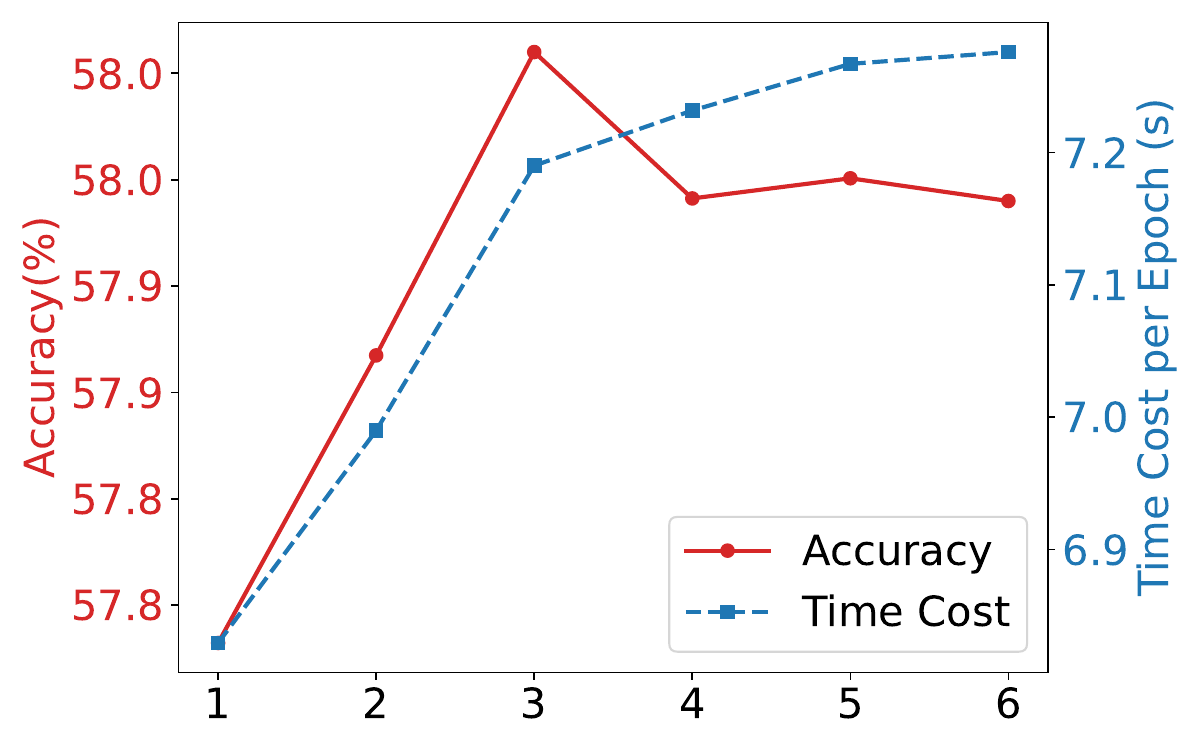}
}
\caption{
Hyper-parameter analysis. 
}
\label{fig:parameter}
\end{figure}
To further validate the model's sensitivity to hyperparameters, we conduct a detailed parameter sensitivity analysis on four core hyperparameters in HOPE using the Ogbn-mag dataset with backbone HGAMLP. Figure~\ref{fig:parameter} illustrates the impact of parameter changes on model performance.

\textbf{Impact of Orthogonality Constraint Weight ($\lambda$)}.
Experimental results show that as $\lambda$ gradually increases from 0.25 to 0.5, model performance exhibits an upward trend, but when $\lambda$ exceeds 0.5, performance begins to decline significantly. This ``inverted U-shaped'' trend reveals a trade-off between diversity and fitting capability. A moderate $\lambda$ is crucial for preventing expert homogenization (Mode Collapse), as it forces experts to learn complementary semantic patterns. However, an excessively large penalty coefficient overly constrains the parameter space of the experts, limiting their flexibility to fit the true data distribution, leading to underfitting.

\textbf{Impact of Similarity Threshold ($\delta$):} 
Model performance peaks within the interval $\delta \in [0.4, 0.8]$, proving that this range is the optimal sweet spot for distinguishing valid signals from noise. When the threshold is too low ($\delta < 0.4$), the model introduces a large number of low-relevance noise connections, which not only dilutes valid features but also increases unnecessary computational burdens; conversely, when the threshold is set too high ($\delta > 0.8$), valid connections with potential semantic associations are erroneously filtered out, leading to model degradation. At this point, the routing mechanism relies overly on mandatory lower-bound allocation, losing its capacity for flexible selection.

\textbf{Impact of Lower Bound Capacity ($K$):} 
For each expert, the minimum number of assigned nodes is $K \times B / N_E$, where $B$ is the batch size, and $N_E$ represents the number of experts. We observe that optimal range of $K$ is $[0.6, 0.7]$. 
This parameter is directly related to the representation quality of long-tail nodes. An overly small lower bound causes some nodes with weak features to fail to obtain sufficient expert resources, becoming ``orphan nodes'' lacking effective gradient updates. While an overly large lower bound is counterproductive, forcing experts to process nodes irrelevant to their semantics, introducing severe interference noise, thereby damaging the specialization of the experts.

\textbf{Impact of Upper Bound Capacity ($C$):} 
For each expert, the maximum number of assigned nodes is $K \times B / N_E$, where $B$ is the batch size, and $N_E$ represents the number of experts.
As $C$ increases from 2 to 4, model performance experiences a sharp rise. The performance improvement in the initial stage indicates that appropriately relaxing the upper limit of expert capacity effectively alleviates the congestion problem caused by ``Hub Nodes'' and fully releases the expressive power of experts. However, when the capacity exceeds 4.0 times the average load, the model reaches a saturation state. At this point, further increasing the capacity upper limit no longer brings performance gains but only increases computational overhead, validating the necessity of finding a balance point between efficiency and performance.

\section{Related Work}

\subsection{Heterogeneous Graph Neural Networks}
HGNNs are now the paradigm for multi-relational graph modeling. Early methods evolved from simple relation-aggregation (e.g., R-GCN~\cite{R-GCN}) to attention-based architectures, such as the hierarchical attention in HAN~\cite{HAN} and the Transformer-driven dynamic modeling in HGT~\cite{HGT}. Subsequent research refined structural awareness via relation-guided attention~\cite{R-GSN} and cross-relation interaction learning~\cite{R-HGNN}. To improve efficiency, approaches like Simple-HGN~\cite{Simple-HGN} and NARS~\cite{NARS} simplified edge embeddings and decoupled neighbor dependencies. More recently, SeHGNN~\cite{SeHGNN} and HGAMLP~\cite{HGAMLP} achieved state-of-the-art scalability by separating feature propagation from transformation and incorporating node-adaptive weighting. Other works have focused on capturing complex topologies, addressing long-range dependencies~\cite{NLA-GNN} and multi-scale hierarchies~\cite{MSHGT}.

Despite these advances, most methods rely on a fixed-capacity global linear decoder. This design overlooks non-linear decision boundaries and fails to handle long-tail distributions, leading to under-fitting on hub nodes and over-smoothing on tail nodes.

\subsection{Mixture-of-Experts}
MoE enables efficient model scaling through conditional computation~\cite{shazeer2017outrageously}. Following foundational sparse gating works like GShard~\cite{lepikhin2021gshard} and Switch Transformer~\cite{fedus2022switch}, research has shifted toward optimizing routing strategies for better load balancing~\cite{zhou2022mixture} and differentiability~\cite{puigcerver2024from}. Additionally, integrating parameter-efficient tuning with experts (e.g., LoRAMoE~\cite{dou2023loramoe}, MoLE~\cite{wumixture}) offers a viable path for resource-constrained training.

\textbf{MoE in Graphs.} Current applications of MoE in graphs primarily enhance the encoder's ability to handle semantic and structural diversity. Methods like GME~\cite{GME}, MoGL~\cite{MoGL}, and MoE-HGT~\cite{MoE-HGT} deploy experts to capture varied connectivity patterns and heterogeneous semantics. Innovations also extend to geometric spaces, with GraphMoRE~\cite{GraphMoRE} utilizing Riemannian experts, and structural learning, where approaches like MoSE~\cite{MOSE} and DA-MoE~\cite{DA-MoE} focus on motif encoding and adaptive receptive fields.

However, existing innovations are restricted to the encoding phase. The decoding stage typically defaults to a static, linear projection. This uniform decision boundary is ill-suited for the complex non-linear patterns and long-tail semantic disparities inherent in heterogeneous graphs.

\section{Conclusion}
To address the ``Fixed Linear Projection Bottleneck'' in HGNNs, we propose HOPE~(Heterogeneous-aware Orthogonal Prototype Experts), a plug-and-play MoE framework. By integrating prototype-based routing, elastic capacity, and orthogonality constraints, HOPE effectively captures diverse and long-tail semantics. Experiments on four datasets confirm that HOPE consistently boosts SOTA performance in node classification and link prediction tasks with minimal overhead, validating the paradigm shift from static linear heads to dynamic expert decoding.

\section*{Impact Statement}
This paper presents work whose goal is to advance the field of 
Machine Learning. There are many potential societal consequences 
of our work, none which we feel must be specifically highlighted here.


\bibliography{example_paper}
\bibliographystyle{icml2025}

\newpage
\appendix
\onecolumn
\section{Link Prediction}
\begin{table*}[t]
\centering
\caption{Link prediction results. OOM denotes Out of Memory, and table shows the time spent in one epoch.
}
\label{tab:experiment-lp}
\fontsize{9pt}{4pt}\selectfont
\setlength{\tabcolsep}{3pt}
{
\begin{tabular}{c|ccccc|ccccc}
\toprule
 & \multicolumn{5}{c|}{Yelp} & \multicolumn{5}{c}{DBLP} \\
\midrule
 & \#Parma & Training & Inferring & AUC & Precision & \#Parma & Training & Inferring & AUC & Precision\\
\midrule[1pt]
R-GCN & 0.11M & 64.90s & 435.38s & \Acc{28.57}{0.17}&  \Acc{43.13}{0.12} & \multicolumn{5}{c}{OOM}\\
\midrule
with HOPE & 0.14M & 63.88s & 437.69s & \Acc{30.74}{0.15}&  \Acc{46.52}{0.18} &\multicolumn{5}{c}{OOM}\\
\midrule[1pt]
R-GAT & 0.11M & 64.78s & 425.73s & \Acc{36.46}{8.62}&  \Acc{44.33}{4.41} & \multicolumn{5}{c}{OOM}\\
\midrule
with HOPE & 0.14M & 67.14s & 448.56s & \Acc{39.02}{6.11}&  \Acc{46.75}{3.26} & \multicolumn{5}{c}{OOM}\\
\midrule[1pt]
R-GSN & \multicolumn{5}{c|}{OOM} & \multicolumn{5}{c}{OOM}\\
\midrule
with HOPE & \multicolumn{5}{c|}{OOM} & \multicolumn{5}{c}{OOM}\\
\midrule[1pt]
NARS & 3.8M & 9.56s & 15.22s & \Acc{75.52}{2.91}&  \Acc{76.08}{2.47} & \multicolumn{5}{c}{OOM}\\
\midrule
with HOPE & 4.0M & 10.24s & 18.33s & \Acc{76.90}{2.64}&  \Acc{78.86}{2.44} & \multicolumn{5}{c}{OOM}\\
\midrule[1pt]
SeHGNN & 6.22M & 4.86s & 5.38s & \Acc{74.69}{1.83}&  \Acc{67.26}{1.57} & 3.03M & 8.49s & 7.85s & \Acc{53.09}{0.26} & \Acc{51.79}{0.15}\\
\midrule
with HOPE& 8.44M & 5.39s & 5.42s & \Acc{77.40}{1.27}&  \Acc{69.49}{1.35} & 3.97M & 9.04s & 7.93s & \Acc{56.11}{0.32} & \Acc{54.58}{0.19}\\
\midrule[1pt]
HGAMLP & 5.83M & 4.23s & 4.47s & \Acc{77.86}{0.66}&  \Acc{72.75}{1.38} & 2.93M & 9.99s & 8.19s & \Acc{52.98}{0.39} & \Acc{51.72}{0.20}\\
\midrule
with HOPE & 8.05M & 4.89s & 4.90s & \Acc{79.69}{0.57}&  \Acc{75.57}{1.06} & 3.87M & 11.02s & 8.98s & \Acc{54.62}{0.28} & \Acc{54.95}{0.31}\\
\bottomrule
\end{tabular}}
\end{table*}

To further validate the capabilities of HOPE, in addition to node classification, we evaluate its performance on the link prediction task using the Yelp and DBLP datasets. We employ Precision and AUC as evaluation metrics. The experimental results are summarized in Table~\ref{tab:experiment-lp}. Similar to the node classification task, HOPE demonstrates strong performance enhancement capabilities in link prediction. After integrating HOPE into HGAMLP, Precision improves by 2.56\% and AUC by 3.12\%. Link prediction typically requires capturing finer-grained semantic relationships than node classification. Baseline models mainly rely on linear projections to aggregate neighborhood information, which often encounters bottlenecks when dealing with complex edge relationships. By introducing multiple orthogonal prototype experts, HOPE decouples features originally compressed in a single linear space into multiple complementary semantic subspaces. The significant improvement in experimental data demonstrates that HOPE successfully enriches node representations, enabling more accurate prediction of potential connections. Link prediction tasks often involve a large number of low-degree long-tail nodes. HOPE's elastic capacity allocation mechanism plays a key role here. By setting a lower bound $K$, it ensures that even sparsely connected nodes are allocated sufficient expert computational resources, avoiding the ``representation collapse" problem common in traditional Top-K routing, thereby improving the overall robustness of predictions.
\end{document}